\documentclass{article}

\usepackage[utf8]{inputenc} 
\usepackage[T1]{fontenc}    
\usepackage{microtype}
\usepackage{graphicx}
\usepackage{subfigure}
\usepackage{booktabs} 
\usepackage{tablefootnote}

\usepackage{appendix}

\usepackage{hyperref}

\usepackage[accepted]{icml2019}

\usepackage{etoolbox}
\makeatletter
\patchcmd\@combinedblfloats{\box\@outputbox}{\unvbox\@outputbox}{}{%
   \errmessage{\noexpand\@combinedblfloats could not be patched}%
}%
 \makeatother
 
\icmltitlerunning{Training Neural Networks with Local Error Signals}

\begin{document}

\twocolumn[
\icmltitle{Training Neural Networks with Local Error Signals}



\icmlsetsymbol{equal}{*}

\begin{icmlauthorlist}
\icmlauthor{Arild Nøkland}{equal,stx}
\icmlauthor{Lars H.~Eidnes}{equal,tn}
\end{icmlauthorlist}

\icmlaffiliation{tn}{Trondheim, Norway}
\icmlaffiliation{stx}{Kongsberg Seatex, Trondheim, Norway}

\icmlcorrespondingauthor{Arild Nøkland}{arild.nokland@gmail.com}
\icmlcorrespondingauthor{Lars H.~Eidnes}{larseidnes@gmail.com}

\icmlkeywords{Machine Learning, ICML}

\vskip 0.3in
]



\printAffiliationsAndNotice{\icmlEqualContribution} 

\begin{abstract}
Supervised training of neural networks for classification is typically performed with a global loss function. The loss function provides a gradient for the output layer, and this gradient is back-propagated to hidden layers to dictate an update direction for the weights. An alternative approach is to train the network with layer-wise loss functions. In this paper we demonstrate, for the first time, that layer-wise training can approach the state-of-the-art on a variety of image datasets. We use single-layer sub-networks and two different supervised loss functions to generate local error signals for the hidden layers, and we show that the combination of these losses help with optimization in the context of local learning. Using local errors could be a step towards more biologically plausible deep learning because the global error does not have to be transported back to hidden layers. A completely backprop free variant outperforms previously reported results among methods aiming for higher biological plausibility. Code is available.\footnote{The code for the experiments is available at \url{https://github.com/anokland/local-loss} }

\end{abstract}

\section{Introduction}

Neural networks for classification are typically trained with a global cross-entropy loss, with the prediction error being back-propagated layer-by-layer from the output layer to hidden layers \cite{Hinton86}. The hidden layer weights cannot be updated before the forward and backward pass has completed. This backward locking prevents parallelization of the weight updates. It also prevents reuse of the memory used to store hidden layer activations. Several methods have been proposed to avoid the backward locking and memory reuse problems \cite{JaderbergCOVGSK17,GomezRUG17}.

Back-propagation of global errors is not biologically plausible for a number of reasons \cite{BengioLBL15}. Several more realistic alternatives have been proposed \cite{Bengio14,LeeZFB15,LillicrapCTA14,Nokland16,ScellierB17}. These methods do not seem to scale to larger and more complicated problems like CIFAR10 and ImageNet \cite{Bartunov18}. 

In this paper, we demonstrate that the backward-locking problem can be avoided by layer-wise training of the hidden layers with locally generated errors. The local loss functions do not depend on a globally generated error, the gradient is not backpropagated to previous layers, and the hidden layer weights can be updated during the forward pass. At the inference stage, the network behaves as a standard network trained with global back-prop. When the weights for a hidden layer have been updated, the gradient and the activations do not have to be kept in memory any more. This alleviates the memory requirements when training deep networks. Although we train all layers at the same time, locally generated errors also enables greedily training layers one-at-a-time, which can reduce the memory footprint even more, and also reduce the training time.

Using local errors could be a step towards more biologically plausible deep learning because the global error does not have to be propagated back to hidden layers. The global target can be projected back to hidden layers instead.

Despite the promise that local loss functions can make training faster, more memory efficient, more parallel and more biologically plausible, layer-wise supervised training has been poorly explored in the literature.

\section{Related work}

\subsection{Local Loss Functions}

Local loss functions have been used to pre-train hidden layers independently of the global loss, and this has in certain cases been shown to improve the performance after fine-tuning using global back-propagation \cite{HintonOT06,Bengio07,SalakhutdinovH09,ErhanBCMVB10,VincentLBM08,PaineKHH14,DongGMYS18}. Local loss functions have been used as an auxillary objective to improve performance \cite{LeeXGZT15,ZhangLL16,SzegedyLJSRAEVR15,WangLTL15a,WestonRMC12}. Using supervised layer-wise loss functions, without fine-tuning, has also been explored previously \cite{Mostafa17,MalachS18,MarquezHN18}. The best reported result on CIFAR-10 is 7.2\% using local classifiers and ensembling, approaching the results of global backprop \cite{BelilovskyEO18}. Our contribution is to show that local classifiers combined with a local similarity matching loss can match global backprop in terms of test error.

Training hidden layers with synthetic gradients is another way to avoid the backward locking problem \cite{JaderbergCOVGSK17}. This method uses local loss functions to train sub-networks to approximate the true gradient. The synthetic gradient modules are trained with an L2 loss to predict the true gradient from the layer above. The input to the module is the hidden layer activation and in some cases, the target vector. The method relates to layer-wise supervised training because the target information is used to train hidden layers. The method differs from our approach because we don't try to approximate a back-propagated gradient, instead we utilize the target vector to create an error signal independently of the layers above.

\subsection{Similarity Measures in Neuroscience}

Similarity measures have been used in the neuroscience field to characterize neural activity patterns. Representational similarity analysis (RSA) measures similarity of representations under different experimental conditions \cite{KriegeskorteMB08}. By comparing the activity associated with each pair of experimental conditions one can obtain a representational dissimilarity matrix (RDM), much like the similarity matrix we use in this work. For instance, RSA performed on recordings from the inferior temporal (IT) cortex in monkeys show that neural responses to images are clustered according to object categories \cite{Kiani07}. The category clusters surprisingly match between humans and monkeys when exposed to the same real-world object images \cite{KriegeskorteMRKBETB08}.


\subsection{Similarity Measures in Machine Learning}

The similarity matching loss function in this paper can be related to previous work in unsupervised clustering and feature learning. Suppose we have $n$ datapoints as the columns of matrix $H=(\mathbf{h}_1,\ldots,\mathbf{h}_n)$. A decomposition of this matrix can be expressed as follows:
\begin{equation} \label{eq:nmf}                                                                                           
\min_{C,G} \left \| H - CG  \right \|^2_F
\end{equation}
If we enforce orthogonality on $C$, requiring $C^TC = I$, the minimization in (\ref{eq:nmf}) implements the subspace version of principal component analysis (PCA). If an L1 penalty is placed on $G$, the minimization performs sparse coding. Under the constraint that the $n$ columns of $G$ are one-hot cluster selectors, solving this minimization finds a k-means clustering of the data, where the $k$ columns of $C$ are the cluster centroids. Under the constraint that $H,C, G\geq0$, the minimization finds a non-negative matrix factorization (NMF) of the data. Each of these methods have been used for unsupervised feature learning in computer vision \cite{CoatesN11,Raina2007,lee1999learning}.

Given some self-similarity measure $S(.)$, consider the objective:
\begin{equation} \label{eq:symnmf}                                              
\min_{G} \left \| S(H) - S(G) \right \|^2_F
\end{equation}
If $S(.)$ measures the euclidian distance between data points, this minimization implements multidimensional scaling (MDS). If we define $S(X) = X^TX$ and constrain $G$ to be $\geq0$, the minimization in (\ref{eq:symnmf}) finds what is called the symmetric NMF \cite{Kuang2012}. If instead of non-negativity we enforce orthogonality, requiring $GG^T = I$, the minimization in (\ref{eq:symnmf}) implements \cite{Ding2005,Kuang2012} the family of methods called spectral clustering \cite{Ng2002spectral}. By choosing different similarity measures $S(.)$ in the first term of (\ref{eq:symnmf}), this minimization can perform different graph clustering objectives, namely ratio association, kernel clustering and normalized cuts \cite{Kuang2012}.

The above are \textit{unsupervised} clustering and feature learning methods. For our purposes in this paper, we use what can be seen as a \textit{supervised} clustering loss, where two data points belong to the same cluster if they have the same label. Given a label matrix $Y=(\mathbf{y_1},\ldots,\mathbf{y_n})$, whose columns are the one-hot encoded labels of the data, we minimize:
\begin{equation}                                            
\min_{\theta} \left \| S(NeuralNet(H; \theta)) - S(Y) \right \|^2_F
\end{equation}
Here the matrix $Y$ is fixed, and instead the parameters $\theta$ of $NeuralNet$ are adjusted to minimize the loss. A straight-forward interpretation of this loss is that it encourages the neural network to learn representations of the data such that distinct classes have distinct representations.

This supervised clustering loss is related to methods like linear disciminative analysis (LDA) \cite{Fisher36}, and neighbourhood component analysis (NCA) \cite{GoldbergerHSS05} because they both utilize label information to perform clustering.

\section{Method}

\begin{figure}[h]
  \includegraphics[width=\linewidth]{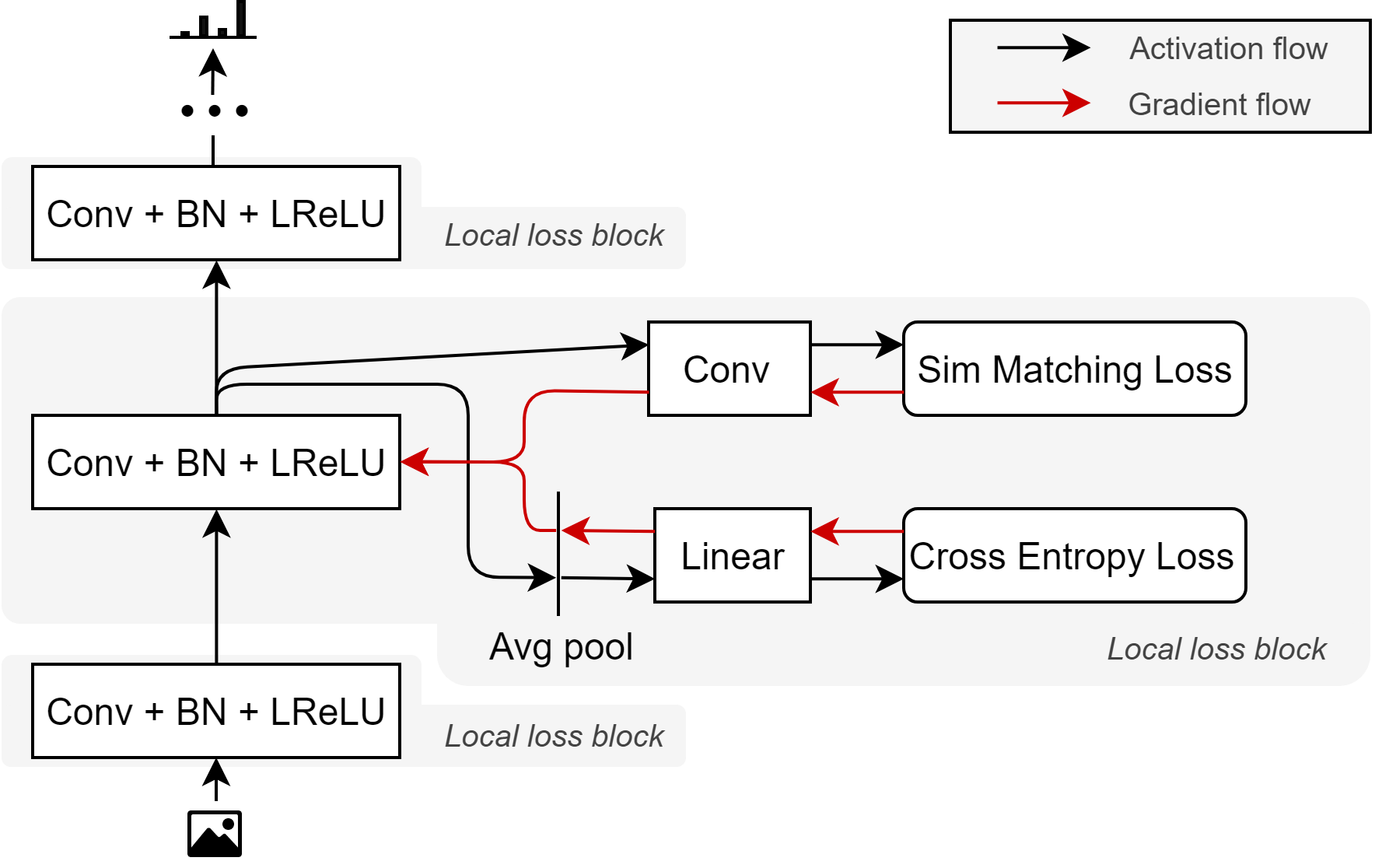}
  \caption{With the local loss functions, each weight layer is trained by two single-layer sub-networks, each with their own distinct loss function.}
  \label{fig:predsimnet}
\end{figure}

We use standard convolutional and fully connected network architectures, but instead of globally back-propagating errors, each weight layer is trained by a local learning signal, that is not back-propagated down the network. The learning signal is provided by two separate single-layer sub-networks, each with their own distinct loss function. One sub-network is trained with a standard cross-entropy loss, and the other with a similarity matching loss (see Figure \ref{fig:predsimnet}).

\subsection{Similarity Matching Loss}
 
The similarity matching loss measures the L2 distance between matrices, where the elements contain the pair-wise similarities between examples in a mini-batch. We denote this loss as $L_{sim}$ or \textbf{sim} loss. Given a mini-batch of hidden layer activations $H=(\mathbf{h}_1,\ldots,\mathbf{h}_n)$, and a one-hot encoded label matrix $Y=(\mathbf{y}_1,\ldots,\mathbf{y}_n)$, we have:
\begin{equation}\label{eq:simloss}                                          
L_{sim} = \left \| S(NeuralNet(H)) - S(Y) \right \|^2_F
\end{equation}
When $H$ is the output of a linear layer, $NeuralNet(.)$ is a linear layer. When $H$ is the output of a convolutional layer, $NeuralNet(.)$ is a convolutional layer with kernel size 3x3, stride 1 and padding 1, followed by a standard deviation operation over each feature map. This operation will reduce the dimension of the output to 2. $S(X)$ is the adjusted cosine similarity matrix, or correlation matrix, of a mini-batch $X$. $S(X)$ contains elements $s_{ij}$ where 
\begin{equation} 
s_{ij} = s_{ji} = \frac{\mathbf{\widetilde{x}}_i^T\mathbf{\widetilde{x}}_j} {\|\mathbf{\widetilde{x}}_i\|_{2} \|\mathbf{\widetilde{x}}_j\|_{2}}
\end{equation}
Subscripts $i$ and $j$ denote indices in the mini-batch. $\mathbf{\widetilde{x}_i}$ denotes the mean-centered vector $\mathbf{x_i}$.

\subsection{Prediction Loss}

The prediction loss measures the cross-entropy between a prediction from a local classifier and the target. We denote this loss as $L_{pred}$ or \textbf{pred} loss.

The \textbf{pred} loss for a matrix $H$ of hidden layer-activations:
\begin{equation} 
L_{pred} = CrossEntropy(Y, W^TH)
\end{equation} 
where $W$ is a weight matrix with width equal to the number of classes and height equal to hidden dimension, and $Y$ is matrix of one-hot encoded targets. If $H$ is the output of a convolutional layer, average-pooling is performed first to reduce the size, then the feature maps are flattened.

\subsection{Backprop Free Version}

A more biologically plausible version of the similarity matching loss is to replace $NeuralNet(.)$ in (\ref{eq:simloss}) with a standard deviation operation over each feature map, and apply the similarity matching objective directly on these features. Then no back-propagation is required to calculate the gradient for the hidden layer. To eliminate the requirement for the global target to be available at each hidden layer, the one-hot encoded target vector is replaced with a random transformation of the same target vector.  We denote this loss as $L_{sim-bio}$ or \textbf{sim-bpf} loss.

The $L_{pred}$ loss can be made more biologically plausible by using feedback alignment \cite{LillicrapCTA14} to transport the prediction error back to the hidden layer. To eliminate the requirement for the global target to be available at each hidden layer, the classifier is trained to predict a binarized random transformation of the target vector using a binary cross-entropy loss. We denote this loss as $L_{pred-bpf}$ or \textbf{pred-bpf} loss.

The experiment section includes one experiment on CIFAR-10 using these two versions, and their combination.

\subsection{Combined Loss}

We denote the weighted combination of the above loss functions as $L_{predsim}$ or simply as \textbf{predsim} loss.
\begin{equation}
L_{predsim} = (1-\beta)L_{pred} + \beta L_{sim}
\end{equation}

And equally for the more biologically plausible loss functions. We denote this loss as $L_{predsim-bpf}$ or simply as \textbf{predsim-bpf} loss.
\begin{equation}
L_{predsim-bpf} = (1-\beta)L_{pred-bpf} + \beta L_{sim-bpf}
\end{equation}

\section{Experiments}

We performed experiments on MNIST, Fashion-MNIST, Kuzushiji-MNIST, CIFAR-10, CIFAR-100, STL-10 and SVHN to evaluate the performance of the training method. We used fully-connected architetures and two VGG-like architectures that we found to perform well \cite{SimonyanZ14a}. For each model we compared the performance when trained with the \textbf{glob} loss (i.e. standard global back-prop), the \textbf{pred} loss, the \textbf{sim} loss and the \textbf{predsim} loss. We also trained the best performing model with cutout regularization \cite{DevriesT17}, keeping all hyper-parameters except the dropout rate identical, to see if it could improve the result further.

The architectures and hyper-parameters were chosen to give good performance for the \textbf{predsim} loss. The hyper-parameters were kept identical for all loss variations for a given dataset and architecture combination. Experiments were kept as simple as possible, and only dropout rate, learning rate, length of training, hidden layer dimension and average-pooling kernel size (used in the \textbf{pred} loss) varied across experiments.

We used two different simple VGG-like convolutional networks. Both consist of 2x2 max-pooling layers, 3x3 convolutional layers with stride 1 and padding 1, and fully connected layers. The first architecture is denoted VGG8B. The layers are \textbf{conv128-conv256-pool-conv256-conv512-pool-conv512-pool-conv512-pool-fc1024-fc}. The dimension of the output layer depends on the number of classes. 

The second network is deeper and is denoted VGG11B. The layers are \textbf{conv128-conv128-conv128-conv256-pool-conv256-conv512-pool-conv512-conv512-pool-conv512-pool-fc1024-fc}. The dimension of the output layer depends on the number of classes.

The experiments were executed using the PyTorch framework. For local loss functions, the computational graph was detached after each hidden layer to prevent backward gradient flow from the output loss. The output layer was trained normally with a cross-entropy loss function. 

A batch size of 128 was used in all experiments. ADAM was used for optimization \cite{KingmaB14}. The weighting factor $\beta$ was manually tuned and set to $0.99$ for all experiments with the \textbf{predsim} loss.

For networks trained with global or \textbf{pred} loss we used the ReLU non-linearity. For networks trained with \textbf{sim} or \textbf{predsim} loss we used leaky-ReLU with a negative slope of 0.01 \cite{Maas13} because it delivered more stable training.  Before each non-linearity we applied batch normalization \cite{IoffeS15}. After each non-linearity we applied dropout \cite{SrivastavaHKSS14}, with equal dropout rate for all layers.

The training time was 100 epochs for MNIST and Kuzushiji-MNIST, 200 epochs for Fashion-MNIST, and SVHN, and 400 epochs for the other datasets. The learning rate was multiplied by a factor of 0.25 at 50\%, 75\% and 89\% and 94\% of the total training time. Because of the high number of experiments, we performed only single-time runs. We report the test error for the last training epoch.

In some of the experiments, the number of convolutional filters were multiplied by a factor of 2 or 3. This is denoted in the tables as (2x) or (3x) trailing the network name. 

Despite the large number pf parameters, we were able to train the networks on a single GPU.

\subsection{MNIST}

MNIST consists of hand-written digits and is the most commonly used dataset within the deep learning community. The dataset is trivial to learn and good performance here does not say much about the performance on harder tasks. We have included these experiments here for completeness. 

The initial learning rate was 5e-4. The average-pooling kernel size for the \textbf{pred} loss was chosen so that the input dimension to the local classifier was 1024. The dropout rate was 0.1 for MLP and 0.2 for VGG8B. For the cutout experiment, the cutout hole size was 14.

We used 2 pixel jittering for data augmentation as done in the CapsNet paper \cite{SabourFH17}. We provide the CapsNet result here as a baseline for convolutional networks, even though better results have been achieved with ensembling and extensive data augmentation \cite{WanZZLF13}.$  $ We provide the performance of the Ladder network as baseline for fully-connected networks \cite{RasmusBHVR15}.

It is clear that jittering is helping substantially for all fully-connected networks, but the best result is with the \textbf{predsim} loss. VGG8B with \textbf{predsim} loss and cutout is on par with CapsNet.

\begin{table}[h]
  \caption{MNIST with 2 pixel jittering. Test error in percent.}
  \label{table:mnist}
  \centering
  \begin{tabular}{lllllll}
    \toprule
    &&& \multicolumn{3}{|l}{Local loss functions} \\
    Model   & \#par & glob & \multicolumn{1}{|l}{pred} & sim & \multicolumn{1}{l}{predsim}  \\
    \midrule
    3x1024 MLP & 2.9M &  0.75 & \multicolumn{1}{|l}{0.68} & 0.80  & \textbf{0.62} \\
    VGG8B      & 7.3M &  \textbf{0.26} & \multicolumn{1}{|l}{0.40} & 0.65  & 0.31 \\
    VGG8B+CO      & 7.3M &  - & \multicolumn{1}{|l}{-} & - & 0.26 \\
    \midrule
    Ladder     & - & 0.57  & \multicolumn{1}{|l}{-} & - & - \\
    CapsNet    & 8.2M & 0.25  & \multicolumn{1}{|l}{-} & - & - \\
    \bottomrule
  \end{tabular}
\end{table}

\subsection{Fashion-MNIST}

Fashion-MNIST is a rather new dataset with different classes of clothing and is a drop-in replacement for MNIST \cite{Fashion17}. It is harder, but has the same size, input dimension and number of classes as MNIST. 

The initial learning rate was 5e-4 for MLP and VGG8B, and 3e-4 for VGG8B(2x). The average-pooling kernel size for the \textbf{pred} loss was chosen so that the input dimension to the local classifier was 1024 for VGG8B and 2048 for VGG8B(2x). The dropout rate was 0.025 for MLP, 0.1 for VGG8B and 0.2 for VGG8b(2x). For the cutout experiment, the cutout hole size was 14.

As a baseline we show the test error for a WideResNet-28-10 \cite{ZagoruykoK16} with and without random erasing data augmentation \cite{ZhongZKLY17}.\footnote{Due to an issue in the early version of the dataset, the baseline numbers reported in \cite{ZhongZKLY17} are too low, see \url{https://github.com/zhunzhong07/Random-Erasing}. We rerun the released code with the current dataset version and report the test error from the last epoch. } This is to our knowledge the best published results on this dataset. Note that the baseline network has about 5 times more parameters than VGG8B. To make the comparison more fair, we also trained a version of VGG8B where the number of convolutional filters were doubled. This version performs better than the baseline, even though the number of parameters was smaller.

\begin{table}[h]
  \caption{Fashion-MNIST with 2 pixel jittering and horizontal flipping. Test error in percent.}
  \label{table:fmnist}
  \centering
  \begin{tabular}{lllllll}
    \toprule
    &&& \multicolumn{3}{|l}{Local loss functions} \\
    Model   & \#par & glob & \multicolumn{1}{|l}{pred} & sim & \multicolumn{1}{l}{predsim}  \\
    \midrule
    3x1024 MLP & 2.9M &  \textbf{8.37} & \multicolumn{1}{|l}{8.60} & 9.70  & 8.54 \\
    VGG8B & 7.3M &  \textbf{4.53} & \multicolumn{1}{|l}{5.66} & 5.12  & 4.65 \\
    VGG8B(2x) & 28M & 4.55 & \multicolumn{1}{|l}{5.11} & 4.92 &  \textbf{4.33} \\
    8B(2x)+CO & 28M & - & \multicolumn{1}{|l}{-} & - &  4.14 \\
    \midrule
    WRN & 37M & 4.63  & \multicolumn{1}{|l}{-} & - & - \\
    WRN+RE& 37M & 4.16  & \multicolumn{1}{|l}{-} & - & - \\
    \bottomrule
  \end{tabular}
\end{table}

\subsection{Kuzushiji-MNIST}

Kuzushiji-MNIST is another drop-in replacement for MNIST containing hand-drawn japanese characters \cite{Kuzushiji18}. 

The initial learning rate was 5e-4. The average-pooling kernel size for the \textbf{pred} loss was chosen so that the input dimension to the local classifier was 1024. The dropout rate was 0.2 for MLP and 0.3 for VGG8B. For the cutout experiment, the dropout rate was 0.15 and the cutout hole size was 14.

As a baseline we have included the first published results on this dataset, a PreActResNet-18 \cite{HeZRS16} with and without manifold mixup regularization \cite{Kuzushiji18}. VGG8B with \textbf{predsim} loss and cutout achieved a test error that surpasses the baseline, even though the number of parameters was smaller.

\begin{table}[h]
  \caption{Kuzushiji-MNIST with no data augmentation. Test error in percent.}
  \label{table:kmnist}
  \centering
  \begin{tabular}{lllllll}
    \toprule
    &&& \multicolumn{3}{|l}{Local loss functions} \\
    Model   & \#par & glob & \multicolumn{1}{|l}{pred} & sim & \multicolumn{1}{l}{predsim}  \\
    \midrule
    3x1024 MLP & 2.9M &  \textbf{5.99} & \multicolumn{1}{|l}{7.26} & 9.80  & 7.33 \\
    VGG8B      & 7.3M &  1.53 & \multicolumn{1}{|l}{2.22} & 2.19  & \textbf{1.36} \\
    VGG8B+CO   & 7.3M &  - & \multicolumn{1}{|l}{-} & -  & 0.99 \\
    \midrule
    PARN       & 11M & 2.18  & \multicolumn{1}{|l}{-} & - & - \\
    PARN+MM    & 11M & 1.17  & \multicolumn{1}{|l}{-} & - & - \\
    \bottomrule
  \end{tabular}
\end{table}

\subsection{CIFAR-10}

CIFAR-10 consist of 50000 training images of dimension 32x32 pixels \cite{Krizhevsky09}. The dataset has 10 classes.

The initial learning rate was 5e-4 for MLP, VGG8B and VGG11B, and 3e-4 for VGG11B(2x) and VGG11B(3x). The average-pooling kernel size for the \textbf{pred} loss was chosen so that the input dimension to the local classifier was 2048 for VGG8B and VGG11B, and 4096 for VGG11(2x) and VGG11B(3x). The dropout rate was 0.1 for MLP, 0.2 for VGG8B and VGG11B, 0.25 for VGG11B(2x) and 0.3 for VGG11B(3x). For the cutout experiment, the cutout hole size was 16.

As a baseline we have included test error for WideResNet-40-10 with and without cutout \cite{DevriesT17}. Note that better results have been reported using regularization and data augmentation techniques  \cite{VermaLBNCMB18,CubukZMVL18}. The \textbf{predsim} loss worked better than the other loss functions for the tested architectures. By multiplying the number of convolutional filters by 3, the VGG11B model trained with \textbf{predsim} loss approaches the test error of WideResNet.

\begin{table}[h]
  \caption{CIFAR10 with standard data augmentation. Test error in percent.}
  \label{table:cifar10}
  \centering
  \begin{tabular}{lllllll}
    \toprule    
    &&& \multicolumn{3}{|l}{Local loss functions} \\
    Model   & \#par & glob & \multicolumn{1}{|l}{pred} & sim & \multicolumn{1}{l}{predsim}  \\
    \midrule
    3x3000 MLP & 27M &  33.6 & \multicolumn{1}{|l}{32.3} & 33.5  & \textbf{30.9} \\
    VGG8B      & 8.9M &  5.99 & \multicolumn{1}{|l}{8.40} & 7.16  & \textbf{5.58} \\
    VGG11B     & 12M &  5.56 & \multicolumn{1}{|l}{8.39} & 6.70  & \textbf{5.30} \\
    VGG11B(2x) & 42M & 4.91 & \multicolumn{1}{|l}{7.30} & 6.66  & \textbf{4.42} \\
    VGG11B(3x) & 91M & 5.02 & \multicolumn{1}{|l}{7.37} & 9.34\tablefootnote{The test error was 5.60\% in epoch 399.} & \textbf{3.97} \\
    11B(3x)+CO & 91M & - & \multicolumn{1}{|l}{-} & - & 3.60 \\
    \midrule
    WRN        & 56M & 3.87  & \multicolumn{1}{|l}{-} & - & - \\
    WRN+CO     & 56M & 3.08  & \multicolumn{1}{|l}{-} & - & - \\
    \bottomrule
  \end{tabular}
\end{table}

We also tested the backprop free training methods on this dataset, with a random target projection of size 128. The weighting factor $\beta$ was set to $0.01$ for the \textbf{predsim-bpf} loss. 

The initial learning rate was 5e-4 for VGG8B, and 3e-4 for VGG8B(2x). The average-pooling kernel size for the \textbf{pred} and \textbf{sim} loss was chosen so that the flattened output dimension was 4096. The dropout rate was 0.05 for VGG8B and 0.1 for VGG8B(2x).

The best published results on this task with no back-propagation is to our knowledge 16.9\% using dense feedback alignment \cite{MoskovitzLKA18}. The second best result is 18.0\% using K-means and SVM \cite{CoatesN11}. If we consider sign-concordant feedback as backprop free training, the best result is 12.6\% \cite{MoskovitzLKA18}. Our result with the \textbf{predsim-bpf} loss surpasses these results by a large margin.

\begin{table}[h]
  \caption{CIFAR10 with standard data augmentation. No back-propagation. Test error in percent.}
  \label{table:cifar10-bio}
  \centering
  \begin{tabular}{lllllll}
    \toprule
    Model       & \#par & pred-bpf & sim-bpf & predsim-bpf  \\
    \midrule
    VGG8B       & 8.9M & 9.80 & 13.39  & \textbf{9.02} \\
    VGG8B(2x)   & 31M & - & -  & 7.80 \\
    \bottomrule
  \end{tabular}
\end{table}

\subsection{CIFAR-100}

CIFAR-100 consist of 50000 training images of dimension 32x32 pixels \cite{Krizhevsky09}. The dataset has 100 classes. 

The initial learning rate was 5e-4 for MLP, VGG8B and VGG11B, and 3e-4 for VGG11B(2x) and VGG11B(3x). The average-pooling kernel size for the \textbf{pred} loss was chosen so that the input dimension to the local classifier was 4096. The dropout rate was 0.025 for MLP, 0.05 for VGG8B and VGG11B, 0.1 for VGG11B(2x) and 0.15 for VGG11B(3x). We were not able to improve the test error with cutout on this dataset, so this result is excluded from the table. 

Because of the high number of classes, we limited the number of classes in each mini-batch to 20 until first drop in learning rate. This was to make the  target similarity matrix $S(.)$ less sparse, and we found that this improved the final result.

As a baseline we have included test error for WideResNet-40-10 with and without cutout \cite{DevriesT17}. Note that better results have been reported using regularization and data augmentation techniques  \cite{YamadaIK18,CubukZMVL18}. The \textbf{predsim} loss worked better than the other loss functions for the tested architectures. By multiplying the number of convolutional filters by 3, the VGG11B model trained with \textbf{predsim} loss approaches the test error of WideResNet.

\begin{table}[h]
  \caption{CIFAR100 with standard data augmentation. Test error in percent.}
  \label{table:cifar100}
  \centering
  \begin{tabular}{lllllll}
    \toprule    
    &&& \multicolumn{3}{|l}{Local loss functions} \\
    Model   & \#par & glob & \multicolumn{1}{|l}{pred} & sim & \multicolumn{1}{l}{predsim}  \\
    \midrule
    3x3000 MLP & 27M &  62.6 & \multicolumn{1}{|l}{58.9} & 62.5  & \textbf{56.9} \\
    VGG8B      & 9.0M &  26.2 & \multicolumn{1}{|l}{29.3} & 32.6 & \textbf{24.1} \\
    VGG11B     & 12M &  25.2 & \multicolumn{1}{|l}{29.6} & 30.8  & \textbf{24.1} \\
    VGG11B(2x) & 42M &  23.4 & \multicolumn{1}{|l}{26.9} & 28.0  & \textbf{21.2} \\
    VGG11B(3x) & 91M & 23.7 & \multicolumn{1}{|l}{25.9} & 28.0 & \textbf{20.1} \\
    \midrule
    WRN        & 56M & 18.8  & \multicolumn{1}{|l}{-} & - & - \\
    WRN+CO     & 56M & 18.4  & \multicolumn{1}{|l}{-} & - & - \\
    \bottomrule
  \end{tabular}
\end{table}

\subsection{SVHN}

SVHN is a dataset with house number images of dimension 32x32 pixels \cite{Netzer2011}. The training set has 73257 images and the extra training set has 531131 images. We used both training sets in our experiments. No data augmentation was used. 

The initial learning rate was 3e-4. The average-pooling kernel size for the \textbf{pred} loss was chosen so that the input dimension to the local classifier was 2048. The dropout rate was 0.3. For the cutout experiment, the dropout rate was 0.15 and the cutout hole size was 16.

As a baseline, we show the test error for WideResNet-16-8 with and wihout cutout \cite{DevriesT17}. Note that better results are reported with extensive data augmentation \cite{CubukZMVL18}. The \textbf{predsim} loss clearly worked better than the other loss functions for the tested architecture, but lags behind the test error for WideResNet. 

\begin{table}[h]
  \caption{SVHN with extra training data, but no data augmentation. Test error in percent.}
  \label{table:svhn}
  \centering
  \begin{tabular}{lllllll}
    \toprule
    &&& \multicolumn{3}{|l}{Local loss functions} \\
    Model   & \#par & glob & \multicolumn{1}{|l}{pred} & sim & \multicolumn{1}{l}{predsim}  \\
    \midrule
    VGG8B & 8.9M &  2.29 & \multicolumn{1}{|l}{2.12} & 1.89  & \textbf{1.74} \\
    VGG8B+CO & 8.9M &  - & \multicolumn{1}{|l}{-} & -  & 1.65 \\
    \midrule
    WRN & 11M & 1.60  & \multicolumn{1}{|l}{-} & - & - \\
    WRN+CO & 11M & 1.30  & \multicolumn{1}{|l}{-} & - & - \\
    \bottomrule
  \end{tabular}
\end{table}

\subsection{STL-10}

STL-10 is a dataset of images belonging to 10 classes \cite{CoatesNL11}. The image dimension is 96x96 pixels. The training dataset consists of 5000 labeled images and a lot of unlabeled images. We used only labeled images, and we did not use the prescribed testing protocol, we just trained one model on all training examples. No data augmentation was used, making this a difficult task because of the small amount of training data. 

The network architecture here was identical to the earlier description, except that the first convolutional layer was replaced with a 7x7 kernel layer with stride 2. This was to reduce the feature map size early in the network. 

The initial learning rate was 5e-4. The average-pooling kernel size for the \textbf{pred} loss was chosen so that the input dimension to the local classifier was 2048. The dropout rate was 0.1. For the cutout experiment, the cutout hole size was 48.

As a baseline we show the results for WideResNet-16-8 with and without cutout regularization \cite{DevriesT17}. The authors use the same training and testing protocol as used here. Our result with \textbf{predsim} loss is better than the baseline.

\begin{table}[h]
  \caption{STL-10 with no data augmentation. Test error in percent.}
  \label{table:stl10}
  \centering
  \begin{tabular}{lllllll}
    \toprule
    &&& \multicolumn{3}{|l}{Local loss functions} \\
    Model   & \#par & glob & \multicolumn{1}{|l}{pred} & sim & \multicolumn{1}{l}{predsim}  \\
    \midrule
    VGG8B & 12M &  33.08 & \multicolumn{1}{|l}{26.83} & 23.15  & \textbf{20.51} \\
    VGG8B+CO & 12M &  - & \multicolumn{1}{|l}{-} & -  & 19.25 \\
    \midrule
    WRN & 11M & 23.48  & \multicolumn{1}{|l}{-} & - & - \\
    WRN+CO & 11M & 20.77  & \multicolumn{1}{|l}{-} & - & - \\
    \bottomrule
  \end{tabular}
\end{table}

\section{Discussion}

\subsection{Results with Local Loss Functions}

Our first observation is that the local prediction loss (\textbf{pred}) achieved test errors close to those of global back-propagation. This is in line with previous work \cite{Mostafa17,BelilovskyEO18}, but still interesting because hidden layers are decoupled from the layers above during training.

A surprising observation is that the local similarity matching loss (\textbf{sim}) is able to provide a remarkable good training signal for hidden layers. In some cases, the test error is lower than for a back-propagated global cross-entropy loss. The loss encourages examples from distinct classes to have distinct representations, measured by the cosine similarity. This can be seen as a kind of supervised clustering. This objective is sufficient to create a hidden representation that is suitable for classification, independently of the layers above.

The results indicate that training with a local cross-entropy (\textbf{pred}) or similarity matching (\textbf{sim}) loss alone does not match a global loss in terms of test error. However, if both loss functions are combined (\textbf{predsim}), the results improve significantly. The performance of the training method varies across architectures and datasets. Overall, we observe no loss in accuracy when comparing to global back-propagation in VGG-like architectures. This conclusion is based on the assumption that the global loss results for VGG architectures are representative. If we compare them with results for residual-free architectures reported in the literature and in open-source implementations, they seem to be equally good or better.

For supervised layer-wise training on CIFAR-10, we improve the state-of-the-art from 7.2\% \cite{BelilovskyEO18} to 3.6\% test error. Layer-wise training of VGG-like models is competitive with residual architectures trained with global back-propagation on several datasets. For STL-10 with no data augmentation, our result is the best reported.

We use dropout \cite{SrivastavaHKSS14}, and batch-normalization \cite{IoffeS15}, for regularization. Without these, the results are much worse. We have demonstrated that more advanced regularization methods like cutout, \cite{DevriesT17}, can improve the results further.

We found that VGG-like architectures work best with the proposed training method. For residual architectures like ResNet and WideResNet, we got better results when the residual connections were removed. We also tried to replace max-pooling layers with 2-strided convolutional layers, but this did not work equally well.

Avoiding a global loss function has several benefits for practical neural network training. The backward-locking problem is no longer a problem, and weights can be updated during the forward pass. This alleviates the memory requirements since activations do not have to be kept in memory for the backward pass. We trained all layers simultaneously, but using local loss functions can also enable greedy training of hidden layers one-by-one. It also allows for model and data parallelism, where different parts of the model can be trained on different GPU's, with each GPU processing different batches.

\subsection{Decoupling Optimization from Generalization}

We have compared test errors of various losses on a wide range of datasets. These results do not on their own help us disentangle effects on optimization from effects on generalization. Looking at the training error can shed light on this. In general, full backprop achieved a faster drop in training error, and as low or lower final training error, compared to the local losses in our experiments. This is not too surprising, considering it has access to the true gradient of the global loss at each layer. 

\begin{figure}[h]
  \includegraphics[width=\linewidth]{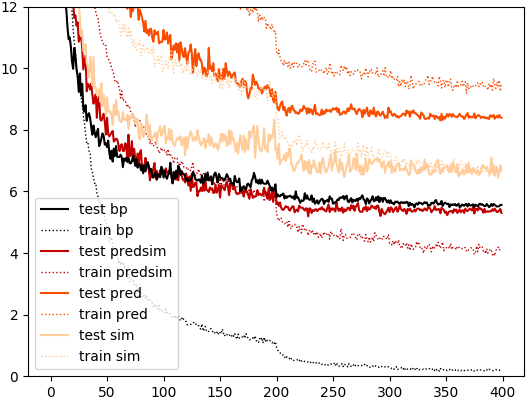}
  \caption{Training and test classification errors on CIFAR10 with full backprop and different local loss functions, with a VGG11B(1x) network.}
  \label{fig:cifar10trainerrors}
\end{figure}

On STL-10, the predsim loss achieved the best reported test error without using unlabeled data. This dataset is characterized by relatively large images (96x96), and few training examples (5000). Large models are prone to overfit on this data. Both backprop and each of the local losses were able to reach a training error < 0.2\%. At the same time, each of the local losses found solutions with lower test error than that of backprop. This immediately suggests that local learning may provide an inductive bias towards solutions that generalize well.

Observing that the combined local loss generally achieves lower test error than each loss on their own, one could argue that the \textbf{sim} loss may be adding a regularizing effect. Looking at the training curves (Figure \ref{fig:cifar10trainerrors}) shows that this is \textit{not} the full story. Here, the \textbf{sim} loss achieved a significantly lower training error than \textbf{pred}, and their combination in \textbf{predsim} achieved lower training error than either method on its own. In every experiment where the local losses could be compared, it held true that the \textbf{predsim} loss achieved a lower training error than either local loss on its own. This shows that both losses help \textit{optimization} in the context of local learning in deep networks.

To summarize, our evidence points to full backprop generally achieving a faster drop in training error, and lower final training error, but that local learning in many experiments appeares to add an inductive bias that reduces overfitting. In the context of local learning, the \textbf{sim} and \textbf{pred} losses both help with optimization in a complementary way. 

We have shown that the \textbf{predsim} loss can achieve strong test error results on many datasets. If our results come from better generalization, one could suspect local learning to perform worse on large datasets like ImageNet, where models are less prone to overfit. Due to time and compute constraints, we do not have ImageNet results in this paper. However, recent results from \cite{BelilovskyEO18} established that local layer-wise training can scale to ImageNet, by achieving a surprisingly low top-5 single-crop test error of 10.2\% on ImageNet with a loss similar to what we call \textbf{pred} loss. This leaves open the possibility that adding a \textbf{sim} loss could improve this result further. 

\subsection{Biological Plausibility}

We have proposed a combination of local loss functions that offers an alternative to end-to-end training with global back-propagation. Neither of the two losses provide backprop free training. However, the error does not have to be transported back through the whole network, a single step of back-propagation is sufficient. 

If we remove the back-propagation requirement and add direct projections from the global target to hidden layers (\textbf{predsim-bpf}), the performance deteriorated, but 7.8\% error on CIFAR-10 is still the best reported result for backprop free methods. This method is biologically plausible in many ways, but some issues still exist. We use unrealistic weight-sharing in convolutional layers, and we allow the weights to switch sign. We also use batch-normalization, which doesn't have a biologically realistic counterpart. The method is an offline algorithm since we use mini-batch training. In addition we have ignored that real neurons communicate with spikes.

Local loss functions could be a step towards a solution to the credit assignment problem \cite{BengioLBL15}. Using local classifiers to generate training signals has been investigated previously. Our contribution is to show that such classifiers can be trained with feedback alignment, and that the target can be replaced with a random projection of the target. We also show increased accuracy when the prediction loss is combined with a similarity matching loss.

With the backprop free \textbf{pred-bpf} loss, the weight transport problem, \cite{Grossberg87}, is avoided because feedback alignment does not  require symmetric weights. An online version of feedback alignment learning can be implemented in a biologically plausible way using multi-compartment neurons \cite{GuerguievLR17}, and in cortical microcircuits for continuous learning without separate forward and backward passes \cite{Sacramento18}. An online version of the \textbf{sim-bpf} loss can be implemented using local Hebbian and anti-Hebbian learning rules, at least for unsupervised learning \cite{Pehlevan2014hebbian,PehlevanSC18, GiovannucciMPC18}. 

Biologically plausible algorithms for error-driven learning have so far focused on how to transport the error back to hidden layers \cite{LillicrapCTA14,Nokland16}, or how to transport a target back to hidden layers \cite{XieS03,Bengio14,LeeZFB15,ScellierB17,WhittingtonB17}. In the context of experiments performed in this paper, global error transportation is not a requirement for error-driven learning. Neither is it a requirement to propagate the global target backwards through the network. The hidden layers can be trained independently of the layers above, without loss in accuracy.

\newpage
\bibliography{local_error}
\bibliographystyle{icml2019}

\newpage
\onecolumn
\appendix
\appendixpage
\section{Training and Test Errors}

\begin{figure*}[h]
  \includegraphics[width=\linewidth]{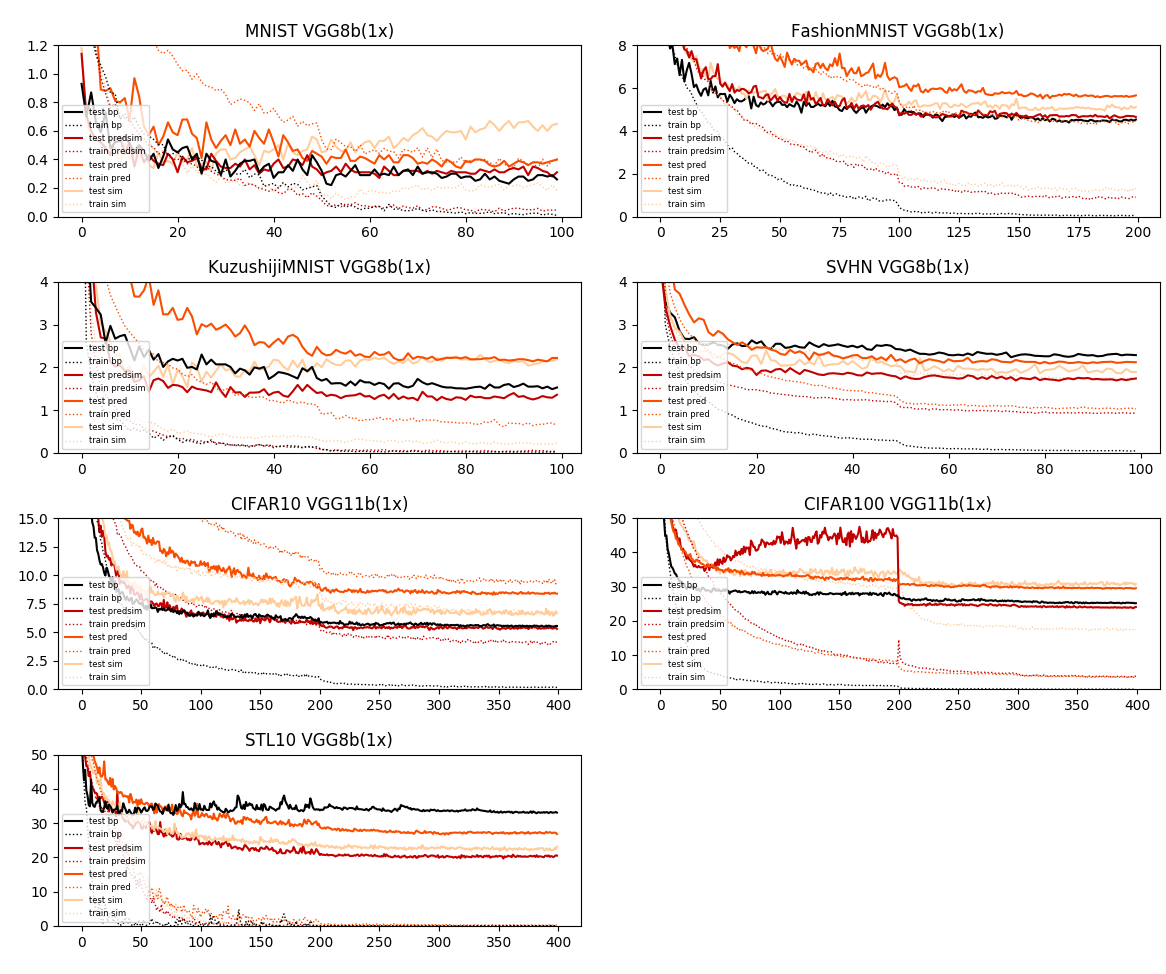}
  \caption{Training and test classification errors on all datasets with different loss functions. Note that the CIFAR100 runs are less comparable to each other, because the sim and predsim runs had batches sampled to have only 20 classes per batch during training, which we found to cause a higher training error, but lower test error. }
  \label{fig:errors_all}
\end{figure*}

\newpage
\section{Similarity Matching as a Complementary Objective}
Although somewhat out of scope of this work, we performed some experiments with the \textbf{sim} loss as a local complementary loss. We trained the networks with global back-propagation combined with a local \textbf{sim} loss. In this way, hidden layers were trained based on a global cross-entropy loss, and back-propagated similarity matching losses from the layers above. Hyper-parameters and training details were identical to the experiments with the local \textbf{sim} loss, except that we did not detach the computational graph.

The results are summarized in Table \ref{table:complementary_sim}. We can see that similarity matching is a powerful auxillary objective for classification, also in a global loss context. For all datasets we can see an improvement in test error compared to global back-propagation alone.

\begin{table}[h]
  \caption{Similarity matching as a complementary objective. Test error in percent.}
  \label{table:complementary_sim}
  \centering
  \begin{tabular}{llllll}
    \toprule
    Dataset   & Model & \#par & glob & predsim & glob+sim  \\
    \midrule
    MNIST           & VGG8B & 7.3M & 0.26  & 0.31 & \textbf{0.24} \\
    Fashion-MNIST   & VGG8B & 7.3M & 4.53  & 4.65 & \textbf{4.16} \\
    Kuzushiji-MNIST & VGG8B & 7.3M & 1.53  & 1.36 & \textbf{1.13} \\
    CIFAR-10        & VGG11B & 12M & 5.56  & 5.30 & \textbf{4.33} \\
    CIFAR-100       & VGG11B & 12M & 25.2  & 24.1 & \textbf{22.2} \\
    SVHN            & VGG8B & 8.9M & 2.29  & \textbf{1.74} & 1.95 \\
    STL-10          & VGG8B & 12M  & 33.1  & \textbf{20.5} & 25.6 \\
    \bottomrule
  \end{tabular}
\end{table}


\end{document}